\documentclass[11pt,a4paper]{article}
\usepackage[hyperref]{acl2021}
\usepackage{times}
\usepackage{latexsym}
\usepackage{booktabs}
\usepackage{multirow}
\usepackage{amsmath}
\usepackage{graphicx}

\usepackage{microtype}

\aclfinalcopy

\title{CO-STAR: Conceptualisation of Stereotypes for Analysis and Reasoning}

\author{Teyun Kwon \\
  Department of Computing \\
  Imperial College London \\
  \texttt{john.kwon20@imperial.ac.uk} \\\And
  Anandha Gopalan \\
  Department of Computing \\
  Imperial College London \\
  \texttt{a.gopalan@imperial.ac.uk} \\}

\date{\today}

\begin{document}

\maketitle

\pagestyle{plain}

\begin{abstract}

\textit{\textbf{Warning:} this paper contains material which may be offensive or upsetting.}

While much of recent work has focused on the detection of hate speech and overtly offensive content, very little research has explored the more subtle but equally harmful language in the form of implied stereotypes. This is a challenging domain, made even more so by the fact that humans often struggle to understand and reason about stereotypes. We build on existing literature and present CO-STAR (\textbf{CO}nceptualisation of \textbf{ST}ereotypes for \textbf{A}nalysis and \textbf{R}easoning), a novel framework which encodes the underlying concepts of implied stereotypes. We also introduce the CO-STAR training data set, which contains just over 12K structured annotations of implied stereotypes and stereotype conceptualisations, and achieve state-of-the-art results after training and manual evaluation. The CO-STAR models are, however, limited in their ability to understand more complex and subtly worded stereotypes, and our research motivates future work in developing models with more sophisticated methods for encoding common-sense knowledge.

\end{abstract}

\section{Introduction}

As more information is made available and digested from online sources such as Twitter and Reddit, it becomes easier for statements containing bias to be propagated and amplified, negatively affecting the social groups involved. Humans naturally seem to involve social bias and stereotypes in order to make judgments efficiently~\cite{Tversky-Judgment}, and the inherent nature of this human characteristic makes it difficult to remove oneself completely from one's personal biases and stereotypical views. If such biases are not addressed, it can result in a subtle and unconscious introduction of stereotypes which not only present a distorted and discriminatory portrayal of the targeted individual or group, but also convey messages which are often damaging and detrimental.

In order to prevent these adverse effects, it remains necessary to understand and become aware of the use of stereotypes in language. However, this is a difficult task even for humans~\cite{Recasens-Linguistic,Pryzant-Automatically}, not only because of the fact that people are often unaware of their own biases~\cite{Blair-Malleability,Bargh-cognitive}, but also due to the subtle and implicit ways in which they are expressed.

We therefore research automated approaches to explain and conceptualise any stereotypes involved. We address this particular task, due to the benefits of detailed statements in explaining why a statement is biased and potentially harmful to certain demographic groups~\cite{Gregor-Explanations,ribeiro-trust}. We also focus on written text, as the close relationship between human cognition and natural language is well-established~\cite{Greenwald-Measuring}.

To this end, we propose a novel framework which we call CO-STAR (\textbf{CO}nceptualisation of \textbf{ST}ereotypes for \textbf{A}nalysis and \textbf{R}easoning), in an effort to improve the ability for language models to analyse and generate stereotypes implied in biased statements. We also introduce the CO-STAR training data set, which contains just over 12K structured implied stereotype annotations and their underlying concepts. We perform ablation studies and find that upon manual evaluation, models trained in accordance with this framework are able to generate more accurate and context-specific stereotypes. We therefore establish state-of-the-art results, but also acknowledge their limited ability to understand more complex and subtly worded stereotypes, and encourage future work to explore more sophisticated methods to encode common-sense knowledge and stereotype conceptualisations.

\section{Related Work}

Research concerning social biases has mostly been centred around mitigating bias in language models~\cite{Bolukbasi-Man,Nadeem-StereoSet,Sun-Mitigating} and reducing their impact on downstream classification tasks~\cite{Blodgett-Language,Hovy-Social}. In terms of detecting linguistic manifestations of bias, recent work has mostly explored detecting overtly hateful language~\cite{Waseem-Hateful,Founta-Twitter,Davidson-Automated}. This, however, is limited in the context and scope of our research, as implicit stereotypes account for much of social bias~\cite{Greenwald-Implicit,Nosek-Implicit,Jurgens-Just} and they are not always hostile in nature~\cite{Eagly-Gender,Glick-Ambivalent,Jha-compliment}. Gender bias has also been studied extensively~\cite{Cryan-Detecting,Hitti-Proposed,Voigt-rtgender,Dinan-Multi-Dimensional,Field-Unsupervised}, but this fails to take into consideration other commonly targeted demographic groups and does not address stereotypes directed at them.

As such, the area of social or demographic biases in the form of stereotypes and generalisations is left relatively underexplored. An exception to this is the work by \citet{Sap-Social}, which is most relevant to our research. They define a formalism with which any social or demographic bias can be analysed and explained, identifying details such as the targeted demographic, the implied statement, and whether the person who projected the biased statement is themself a member of the targeted social group or not. They then perform social bias detection and implication generation tasks using their framework and OpenAI's GPT and GPT-2~\cite{Radford-gpt,Radford-gpt2} language models. They are able to perform the classification tasks well, but there seems to be some room for improvement with respect to explaining why a particular statement is biased: the models are able to perform well when the posts are more explicit with the stereotypes they are trying to convey, but less well when the stereotypes are more subtle and the posts do not contain any lexical overlap with the reference sentences, resulting in stereotypes which are too generic and irrelevant to the input text.

We therefore decide to contribute to this nascent field of research. We believe that existing work on hate speech and gender bias, while important in its own right, is too limited in scope. The work of \citet{Sap-Social} aligns closely to our aims, but presents performance limitations. Further, although numerous frameworks to capture common-sense knowledge have been proposed~\cite{rashkin-connotation,Sap-ATOMIC,Bosselut-COMET}, no existing work, to the best of our knowledge, attempts to encode this information when reasoning about implied stereotypes. We hope that our experiments demonstrate the clear research merits of this approach, and encourage further work in this exciting and promising area of research.

\section{The CO-STAR Framework} \label{chapter-improved-stereotype-generation}

CO-STAR is a novel framework which aims to encode information about the core concept of a post's implied stereotype (in other words, what the post or stereotype is about), as well as to formalise it in a structured manner. We hypothesise that the models will be able to generate more accurate outputs with the extra information available during training, making use of the connections learnt between the post and its associated concept. A more structured set of training instances also allows for consistency which may aid the models in learning the desired patterns in the training data set and focusing on the most relevant input tokens.

\subsection{ConceptNet} \label{co-star-conceptnet}

In structuring the implied stereotype, we take inspiration from ConceptNet 5~\cite{Speer-ConceptNet} and more specifically, its relations, which encode common and intuitive patterns between various entities. We structure the implied stereotypes in the following manner:

\begin{center}
    \texttt{[TARGETED GROUP]} \\
    \texttt{[RELATION]} \\
    \texttt{[IMPLIED STATEMENT]}
\end{center}
\noindent
where \texttt{[TARGETED GROUP]} is the social group or demographic targeted in the post, \texttt{[RELATION]} is one of the linking verbs inspired by the relations in ConceptNet 5, and \texttt{[IMPLIED STATEMENT]} completes the rest of the stereotype. For example, \textit{Korean folks have weird names} would be correctly structured, with \textit{Korean folks} being the \texttt{[TARGETED GROUP]}, \textit{have} being the \texttt{[RELATION]}, and \textit{weird names} being the \texttt{[IMPLIED STATEMENT]}; on the other hand, \textit{trivialises harm to victims} would not be following the above structure.

We limit the \texttt{[RELATION]} component to a choice of eight, in order to cover as many semantic possibilities as possible, but also in an effort to restrict the output generation space for easier pattern recognition and generation. We also ensure that both verbal and adjectival phrases can be introduced for the \texttt{[IMPLIED STATEMENT]} while also making grammatical sense with at least one of the choices for the \texttt{[RELATION]}. This is crucial, since stereotypes often manifest in the behaviours of the targeted groups and the attributes associated with them~\cite{Fast-Shirtless}. Two of the eight choices for the \texttt{[RELATION]}, namely \textit{should} and \textit{do}, have not been derived from ConceptNet 5, but are nevertheless relevant in formulating stereotypes. The eight choices for the \texttt{[RELATION]} component are presented in Table~\ref{relations}, along with their ConceptNet counterparts where available.

\begin{table}
    \centering
    \begin{tabular}{ll}
    \toprule
    \texttt{[RELATION]} & ConceptNet\\
    \midrule
    are & /r/RelatedTo\\
    have & /r/HasA\\
    can & /r/CapableOf\\
    cause & /r/Causes\\
    prevent & /r/ObstructedBy\\
    want & /r/Desires\\
    should & \multicolumn{1}{c}{--}\\
    do & \multicolumn{1}{c}{--}\\
    \bottomrule
    \end{tabular}
    \caption{Relations used as part of the annotation procedure, with their approximate ConceptNet 5 counterparts.}
    \label{relations}
\end{table}

\subsection{Stereotype Conceptualisation} \label{co-star-stereotype-conceptualisation}

As part of the training data set, we also include the core concept of what each post and its implied stereotype is about. This would provide extra information as to which parts of the post are more important, allowing the model to generate more context-specific stereotypes. We present example posts and their corresponding stereotype conceptualisations in Table~\ref{stereotype-conceptualisation-examples}.

We ensure that the conceptualisations are kept short, and limit the length to a maximum of three words. The reason for this is to allow the model to learn what the stereotype is about without having to focus on too many tokens. In addition, we require these to be worded differently to the \texttt{[TARGETED GROUP] [RELATION] [IMPLIED STATEMENT]} tuple when the posts in the training data set are annotated. This is to ensure that as much relevant information as possible is provided to the models during training.

\begin{table*}
    \centering
    \begin{tabular}{lp{0.6\linewidth}l}
    \toprule
    & post & stereotype conceptualisation\\
    \midrule
    (a) & \textit{All-female casts bother me but I have literally never noticed the hundreds of movies with 0 female characters I have already watched} & actor gender preference\\
    (b) & \textit{As expected, when the terrorist group Hamas won the election in Gaza it was the usual one man, one vote, one time, Islamist election. \#Islam} & authoritarian government\\
    (c) & \textit{Black guy in class: *attempts to throw a paper ball into the trash* *misses* Mrs. Mundy: ``You're a disgrace to your race, Marcus''} & athletic skill\\
    (d) & \textit{How can a Libtard speak truth to power when the only thing they vomit from their mouths is lies?} & incompetent policymaking\\
    (e) & \textit{I hate fat bitches} & physical appearance\\
    \bottomrule
    \end{tabular}
    \caption{Examples of stereotype conceptualisations.}
    \label{stereotype-conceptualisation-examples}
\end{table*}

\section{Data Annotation}

We annotate the training data set in accordance with the CO-STAR framework, and conduct data annotation on the Amazon Mechanical Turk platform.\footnote{\texttt{\href{https://www.mturk.com/}{https://www.mturk.com/}}} For each post, annotators provide one stereotype implied by the post in the form of the \texttt{[TARGETED GROUP] [RELATION] [IMPLIED STATEMENT]} tuple, and the stereotype conceptualisation in three or fewer words and without simply repeating their answer for the implied stereotype. The \texttt{[RELATION]} field is provided as a drop-down list, and the others as free-text inputs. Some of the most common responses for the \texttt{[TARGETED GROUP]}, \texttt{[IMPLIED STATEMENT]} and stereotype conceptualisation components of the framework are displayed in Table~\ref{conceptualisation}. A screenshot of the full annotation task is shown in the appendix, in Figure~\ref{annotation-screenshot-1}.

\begin{table}[t]
    \centering

    \begin{tabular}{p{0.7\linewidth}r}
    \toprule
    \texttt{[TARGETED GROUP]} & count\\
    \midrule
    women & 2,318\\
    black folks & 1,914\\
    jewish folks & 668\\
    muslim folks & 341\\
    children & 249\\
    men & 214\\
    gay folks & 175\\
    white folks & 169\\
    \bottomrule
    \\
    \end{tabular}

    \begin{tabular}{p{0.7\linewidth}r}
    \toprule
    \texttt{[IMPLIED STATEMENT]} & count\\
    \midrule
    marginalized for a joke & 1,010\\
    sex objects & 150\\
    inferior & 140\\
    bitches & 106\\
    hoes & 90\\
    terrorists & 73\\
    stupid & 70\\
    criminals & 67\\
    \bottomrule
    \\
    \end{tabular}

    \begin{tabular}{p{0.7\linewidth}r}
    \toprule
    stereotype conceptualisation & count\\
    \midrule
    racial hierarchy & 584\\
    name-calling & 515\\
    racism & 430\\
    sexism & 266\\
    gender hierarchy & 224\\
    paedophilia & 170\\
    terrorism & 160\\
    holocaust & 152\\
    \bottomrule
    \end{tabular}
    \caption{Examples of common responses to the \texttt{[TARGETED GROUP]}, \texttt{[IMPLIED STATEMENT]} and stereotype conceptualisation annotation tasks.}
    \label{conceptualisation}
\end{table}

\subsection{Data Set}

We use a subset of the data set gathered by \citet{Sap-Social} for training and evaluation. It includes both overt cases of stereotypical posts, as well as more subtly biased statements in the form of microaggressions. The corpus by \citet{Breitfeller-Microaggressions} contains the latter, and the former are posts taken from tweets released by \citet{Founta-Twitter}, \citet{Davidson-Automated}, and \citet{Waseem-Hateful} which contain instances of toxic or abusive language, and from hate communities known to engage in white-supremacist, neo-Nazi and violently misogynistic ideologies, namely Stormfront~\cite{Gibert-Hate}, Gab\footnote{\texttt{\href{https://files.pushshift.io/gab/GABPOSTS\_CORPUS.xz}{https://files.pushshift.io/gab/
GABPOSTS\_CORPUS.xz}}} and banned subreddits \texttt{r/Incels} and \texttt{r/MensRights}. The full breakdown of the number of posts from each category is presented in Table~\ref{SBICv2}.

\begin{table}[t]
    \centering
    \begin{tabular}{llr}
    \toprule
    type & source & \# posts\\
    \midrule
    \multirow{5}{*}{Reddit} & \texttt{r/DarkJokes} & 5,176\\
                            & \texttt{r/MeanJokes} & 1,732\\
                            & \texttt{r/OffensiveJokes} & 195\\
                            & \citet{Breitfeller-Microaggressions} & 657\\
                            \cmidrule{2-3}
                            & \textit{\textbf{subtotal}} & 7,760\\
    \midrule
    \multirow{4}{*}{Twitter}& \citet{Founta-Twitter} & 658\\
                            & \citet{Davidson-Automated} & 1,124\\
                            & \citet{Waseem-Hateful} & 297\\
                            \cmidrule{2-3}
                            & \textit{\textbf{subtotal}} & 2,079\\
    \midrule
    \multirow{4}{*}{\shortstack{Hate \\ Sites}} & Stormfront & 785\\
                                & Gab & 604\\
                                & Banned Subreddits & 847\\
                                \cmidrule{2-3}
                                & \textit{\textbf{subtotal}} & 2,236\\
    \midrule\midrule
    \multicolumn{2}{l}{\textbf{CO-STAR total \# posts}} & 12,075\\
    \bottomrule
    \end{tabular}
    \caption{The CO-STAR training data set.}
    \label{SBICv2}
\end{table}

\subsection{Annotator Demographics}

We restrict our annotator pool to the U.S. and U.K., and count the total number of unique annotators for the full training data set after clean-up to be 754, averaging just over 16 posts per annotator. As part of the survey, we also provide an option for annotators to provide basic demographic information. Of those who reported the information, we find that the annotators were relatively balanced in gender and age, but skewed in race (Table~\ref{annotator-demographics}).

\begin{table}[t]
    \centering
    \begin{tabular}{ccc}

    \begin{tabular}{lr}
    \toprule
    gender & \%\\
    \midrule
    female & 50\\
    male & 48\\
    other & 2\\
    \bottomrule
    \end{tabular}

    \enspace

    \begin{tabular}{lr}
    \toprule
    race & \%\\
    \midrule
    White & 76\\
    Asian & 8\\
    mixed & 7\\
    Black & 6\\
    other & 3\\
    \bottomrule
    \end{tabular}

    \enspace

    \begin{tabular}{lr}
    \toprule
    age & \%\\
    \midrule
    18-24 & 15\\
    25-34 & 42\\
    35-44 & 24\\
    45-54 & 12\\
    55+ & 7\\
    \bottomrule
    \end{tabular}

    \end{tabular}
    \caption{Reported annotator demographic information.}
    \label{annotator-demographics}
\end{table}

\section{Training} \label{section-improved-generation-training}

We explore two different input concatenation methods in order to analyse their effects. Firstly, each post and its corresponding labels are concatenated as follows:

\begin{align} \label{improved-generation-model-input-1}         \nonumber
    \mathbf{x} = \{&p_1, \dots, p_m, \texttt{[SEP]},\\ \nonumber
    &c_1, \dots, c_n, \texttt{[SEP]},\\
    &t_1, \dots, t_o, r_1, \dots, r_p, s_1, \dots, s_q, \texttt{[EOS]}\}
\end{align}
\noindent
where $p_{1:m}$ is the sequence of tokens for a given post, $c_{1:n}$ is the sequence of tokens for the corresponding stereotype conceptualisation, and $t_{1:o}$, $r_{1:p}$ and $s_{1:q}$ are token sequences for \texttt{[TARGETED GROUP]}, \texttt{[RELATION]} and \texttt{[IMPLIED STATEMENT]} respectively. The \texttt{[SEP]} and \texttt{[EOS]} tokens are used to separate and end the model input respectively.

Secondly, the implied stereotype is positioned before the stereotype conceptualisation:

\begin{align} \label{improved-generation-model-input-2}         \nonumber
    \mathbf{x} = \{&p_1, \dots, p_m, \texttt{[SEP]},\\ \nonumber
    &t_1, \dots, t_o, r_1, \dots, r_p, s_1, \dots, s_q, \texttt{[SEP]},\\
    &c_1, \dots, c_n, \texttt{[EOS]}\}
\end{align}
\noindent
These training methods are similar to those employed by \citet{Bosselut-COMET} for their common-sense generation task. The concatenated training instances are then tokenised, truncated and shuffled. For evaluation, only the post and the subsequent \texttt{[SEP]} token are encoded and passed in as input.

\subsection{Ablation Studies}

In order to understand the effects of the CO-STAR framework more clearly, we perform ablation studies by training models without annotations for the stereotype conceptualisation. Thus, we concatenate the model inputs as follows:

\begin{align} \label{improved-generation-model-input-ablation} \nonumber
    \mathbf{x} = \{&p_1, \dots, p_m, \texttt{[SEP]},\\
    &t_1, \dots, t_o, r_1, \dots, r_p, s_1, \dots, s_q, \texttt{[EOS]}\}
\end{align}

In order to dispel any doubt relating to the model performances, we train two instances of the ablated model, and in cases where one ablated model generated less accurate outputs than the models trained with both implied stereotype and stereotype conceptualisation annotations, we compare their outputs to the other ablated model. This would increase our confidence in the overall performance of the models trained in accordance with the CO-STAR framework, as well as help eliminate any bias against the ablated models and ensure that they are not considered to have underperformed due to any randomness in the training data set.

\subsection{Other Training Details}

All experiments are performed using the Hugging Face Transformers library.\footnote{\texttt{\href{https://github.com/huggingface/transformers}{https://github.com/huggingface/
transformers}}} We use OpenAI's GPT-2 language model~\cite{Radford-gpt2}, which has yielded successful results in common-sense generation and conditional generation tasks~\cite{Bosselut-COMET,Keskar-CTRL}.

We generate three candidate sentences for each post, with a maximum of 50 tokens each. The models are evaluated after each training epoch, and we train the final models for up to 5 epochs with a batch size of 1, and use the Adam optimiser~\cite{Adam-Kingma} with a learning rate of $1 \times 10^{-5}$.

\section{Results} \label{section-improved-generation-results}

Due to the limitations of automatic evaluation metrics especially for generation systems~\cite{Liu-Evaluate}, we perform manual evaluation for analysing the performance of the different models. We randomly sample 500 posts from the development set (out of a total of 1,806 posts) and compare the outputs of the ablated and non-ablated models over all five training epochs to the reference sentences. In reporting the results, we reproduce the posts and their corresponding outputs for each model as they are, including any errors in spelling and grammar. With respect to the names of our trained models, -cs, -sc and -s refer to models trained with data points as per Equations~\ref{improved-generation-model-input-1}, \ref{improved-generation-model-input-2} and~\ref{improved-generation-model-input-ablation}, respectively.

\subsection{Main Findings} \label{improved-generation-results-main-obs}

Upon close evaluation of the 500 posts from the development set, we find that most of the outputs are similar in content across the different trained models. We therefore focus our analysis on only the posts and their corresponding model outputs where there is a significant difference between the generated candidate sentences. We also report the outputs of a model trained from the work by \citet{Sap-Social}, namely SBF-GPT\textsubscript{2}, for comparison. We reproduce the model outputs in Tables~\ref{table-improved-generation-results-1} and~\ref{table-improved-generation-results-2}.

We make five key observations. Firstly, we notice a general trend of models which have been trained with both stereotype conceptualisation and implied stereotype annotations performing better than the ablated models. This is shown by posts (a) and (b) in Table~\ref{table-improved-generation-results-1}. For post (a), the -cs and -sc models are, for the most part, able to generate the correct stereotype that \textit{men need to act manly}, and these outputs are also accompanied by accurate stereotype conceptualisations (\textit{gender norms} and \textit{gender roles}). Exceptions are \textit{men are shallow} and \textit{men are stupid} which are also present in the outputs of CO-STAR-s. Post (b) again shows the effects of the CO-STAR framework, as the -cs model is able to generate accurately the events entailed by the post, namely the atomic bombings of Hiroshima and Nagasaki (the CO-STAR-s model's \textit{japanese folks have been killed in the holocaust} may, however, also be considered to be technically correct), despite the distinct lack of lexical overlap.

\begin{table*}[ht!]
    \centering
    \begin{tabular}{lp{0.8\linewidth}}
    \toprule
    post (a) & \textit{That's not what a real man would do.}\\
    \midrule
    CO-STAR-cs & sexism \texttt{[SEP]} men are shallow / gender roles \texttt{[SEP]} men should be men / gender roles \texttt{[SEP]} men should be masculine\\
    CO-STAR-sc & men are bad at following norms \texttt{[SEP]} gender norms / men are bad at following norms \texttt{[SEP]} sexism / men are stupid \texttt{[SEP]} mental abilities\\
    CO-STAR-s & men are stupid / men are shallow / men should not do anything\\
    SBF-GPT\textsubscript{2} & men are stupid / men don't deserve rights / men don't deserve respect\\
    references & men need to act manly\\
    \toprule
    post (b) & \textit{Why do Japanese people never play bingo? They all disappear after they hear B-29}\\
    \midrule
    CO-STAR-cs & japanese hate planes \texttt{[SEP]} hiroshima and nagasaki victims are marginalized for a joke / japanese hate planes \texttt{[SEP]} hiroshima and nagasaki are marginalized for a joke / japanese military \texttt{[SEP]} nagasaki victims are marginalized for a joke\\
    CO-STAR-sc & japanese folks are marginalized for a joke \texttt{[SEP]} war / japanese folks are marginalized for a joke \texttt{[SEP]} bombing / japanese folks are marginalized for a joke \texttt{[SEP]} war crimes\\
    CO-STAR-s & japanese folks are marginalized for a joke / japanese folks have been killed in action / japanese folks have been killed in the holocaust\\
    SBF-GPT\textsubscript{2} & japanese people are suicide bombers / japanese people are suicide bombers / japanese people are always dying\\
    references & trivializes harm to victims / japanese got bombed to end ww2 / japanese offended by nuke jokes\\
    \toprule
    post (c) & \textit{If your angry punch a black kid What they gonna do tell their fathers?}\\
    \midrule
    CO-STAR-cs & domestic violence \texttt{[SEP]} black fathers are abusive to their children / domestic violence \texttt{[SEP]} black fathers are violent to their children / domestic violence \texttt{[SEP]} black fathers are abusive\\
    CO-STAR-sc & black folks are inferior to other races \texttt{[SEP]} racial hierarchy / black folks are inferior \texttt{[SEP]} racial hierarchy / black folks are inferior to whites \texttt{[SEP]} racial hierarchy\\
    CO-STAR-s & black fathers are not able to provide for their children / black fathers are not present in their children's lives / black fathers do not have fathers\\
    SBF-GPT\textsubscript{2} & black people are violent / black folks are violent / black people are worthless\\
    references & dad's abandon them / no one cares for them / black fathers abandon their kids / blacks come from single mother homes / black people don't have fathers / black men do not provide for their family\\
    \bottomrule
    \end{tabular}
    \caption{A selection of model predictions and reference sentences.}
    \label{table-improved-generation-results-1}
\end{table*}

We attribute the performance improvements to the addition of the stereotype conceptualisation as part of the training data set, as models are able to learn extra associations and connections between words which would otherwise not have been available, and generate relevant and context-specific stereotypes entailed by the post as opposed to the completely incorrect outputs generated by the ablated models. This is further evidenced by the -cs and -sc models accurately generating the stereotype conceptualisations themselves: \textit{gender norms} and \textit{gender roles} for post (a) as stated above, and \textit{war}, \textit{bombing} and \textit{war crimes} for post (b).

Their ability to generate relevant stereotype conceptualisations raises another important point in favour of the CO-STAR framework, as these models are able to provide more information with respect to the content of the post, even when the implied stereotype generations are rather generic (such as \textit{are marginalized for a joke}). For post (b), the extra details generated by the CO-STAR-sc model after \textit{japanese folks are marginalized for a joke} make the overall outputs more context-specific and potentially allow for a better and clearer understanding of not only the inner workings of the different models, but also the contents of the posts for the end user. The ablated models naturally do not provide such information.

The second key observation we make is that despite the apparent benefits of training the models on the CO-STAR framework, the ablated models do occasionally perform better than the -cs and -sc models. Post (c) provides an example of this, as the outputs of the CO-STAR-s model are able to capture almost perfectly the implications entailed by the post (with the possible exception of \textit{black fathers do not have fathers}), whereas the -cs model generated candidates which share similar words as the post but are irrelevant (\textit{black fathers are abusive/violent to their children}), and the -sc model generated vague stereotypes (\textit{black folks are inferior}). This may be due to the over-reliance of the -cs model on the generated stereotype conceptualisations, with the subsequent implied stereotypes being affected by the wrongly predicted \textit{domestic violence}, as the model tries to make sense of it given the context of the post. As for the -sc model, it may simply have been unable to make sense of the post, and defaulted to generic stereotypes.

The third key observation is related to the point about the -cs model relying too much on the generated stereotype conceptualisations, as mentioned briefly above. There are some performance differences between the -cs and -sc models themselves, and the order in which the components of the CO-STAR framework are concatenated seems to be of significance. For example, post (d) in Table~\ref{table-improved-generation-results-2} shows a clear contrast between the generated implied stereotypes of the -cs and -sc models. The former is unable to generate the entailed implications, whereas the outputs of the latter and of the ablated models are almost identical (with \textit{women are incapable of controlling themselves} being generated by both -sc and -s models as one of the three candidate sentences). This may be due to the unidirectional nature of the GPT-2 language model: since the prediction of the next output token is dependent on the previously generated tokens, the presence of the previously generated stereotype conceptualisations may affect the model performance in generating the subsequent implied stereotypes. It may be that the -cs model becomes less focused on the content of the posts as a result, and instead generates stereotypes which commonly follow the predicted stereotype conceptualisations in the training data set. This results in the model generating stereotypes which are less accurate and context-specific (\textit{women are dependent on men}). The -sc model, however, is not prone to this problem, as the implied stereotypes are conditioned directly on the post itself (with the stereotype conceptualisations predicted at the end), and the generated candidate sentences are therefore similar to those of the ablated models.

\begin{table*}[ht!]
    \centering
    \begin{tabular}{lp{0.8\linewidth}}
    \toprule
    post (d) & \textit{Women have as much self control in the presence of Chad trying to fuck them as we would in the presence of a hot woman trying to fuck us. Zero.}\\
    \midrule
    CO-STAR-cs & gender hierarchy \texttt{[SEP]} women are dependent on men / sexism \texttt{[SEP]} women are subservient to men / gender hierarchy \texttt{[SEP]} women are dependent on men for sex\\
    CO-STAR-sc & women are incapable of controlling themselves \texttt{[SEP]} sexism / women are incapable of controlling their sexuality \texttt{[SEP]} sexism / women are incapable of controlling their sexual urges \texttt{[SEP]} sexism\\
    CO-STAR-s & women should be submissive / women are incapable of controlling themselves / women are defined by their sexuality\\
    SBF-GPT\textsubscript{2} & women are sex objects / women are submissive / women can't be trusted\\
    references & women can't resist sex / women have no control / women don't choose sexual partners well / women are biased on looks when it comes to finding a mate / women are hypocritical about their sexual desire / women can't control themselves with attractive men\\
    \toprule
    post (e) & \textit{What works better than a pacifier and is more convenient? My penis}\\
    \midrule
    CO-STAR-cs & pedophilia \texttt{[SEP]} children are sex objects / sexual assault \texttt{[SEP]} rape victims are marginalized for a joke / pedophilia \texttt{[SEP]} victims of pedophilia are made fun of\\
    CO-STAR-sc & men should be beaten \texttt{[SEP]} domestic violence / men are inferior to women \texttt{[SEP]} gender hierarchy / men should have sex with their penises \texttt{[SEP]} sexual activity\\
    CO-STAR-s & men are better off dead / men are better spanking their wives / men want sex with pacifiers\\
    SBF-GPT\textsubscript{2} & women are sex objects / makes light of rape / trivializes rape\\
    references & advocates child molestation / makes light of child abuse\\
    \bottomrule
    \end{tabular}
    \caption{A selection of model predictions and reference sentences.}
    \label{table-improved-generation-results-2}
\end{table*}

Therefore, whenever the ablated models perform well, the performance of the -sc model is often matched and that of the -cs model is not. However, the opposite is also true, albeit occurring less frequently, and this is our fourth key observation: as shown in post (e) in Table~\ref{table-improved-generation-results-2}, the -sc and ablated models underperform, whereas the -cs model is able to predict the correct stereotype conceptualisations (\textit{paedophilia} and \textit{sexual assault}) and the implied stereotypes which follow. The contrast between the outputs of the -cs model and of the -sc and ablated models is still present, and even when the latter models generate less accurate candidate sentences, their contents are still similar (\textit{male sexual activity} and \textit{violence towards men}). This further emphasises the effects and benefits of the stereotype conceptualisation component of the CO-STAR framework, since if the -cs model is able to generate accurate predictions of this component, it is also able to generate the subsequent implied stereotypes just as accurately. It also reinforces the similarity of the -sc and ablated models, and the significance of the order in which the components of the CO-STAR framework are concatenated.

Finally, we observe that the SBF-GPT\textsubscript{2} model, which was the state-of-the-art, is consistently outperformed by at least one of the CO-STAR models. This is despite the fact that the SBF-GPT\textsubscript{2} model was trained on a data set which contained nearly four times as many unique annotations. This again highlights the importance of enabling the models to learn the underlying concepts for each post, allowing them to generate implied stereotypes which are more specific and relevant. The performance improvements of the ablated models compared to the SBF-GPT\textsubscript{2} model also demonstrate the benefits of structuring the implied stereotypes, as this was the only major difference between the training data sets for these models.

\subsection{Limitations}

Despite the superior model performances of the three CO-STAR models, they are nevertheless not perfect, as they are still unable to understand the implications entailed by more complex posts. We suspect that the information encoded in the stereotype conceptualisations is insufficient for posts where the implications are very subtle and specific, and encourage future work to develop more sophisticated models in addition to the CO-STAR framework.

The evaluation method employed in studying the model performances should also be placed under scrutiny. Firstly, the models are evaluated on only a subset of the development set, which may not be fully representative of their overall performances. Secondly, as with any manual evaluation method, there would no doubt be bias related to the criteria used to determine whether one set of model outputs is more accurate or context-specific than another. These issues may be mitigated by undertaking a more thorough and complete evaluation of the model performances via crowdsourcing and obtaining evaluation agreements among the workers, which would not only enable model evaluation for more posts, but also a smoothing out of any bias which may be present in determining the more accurate or context-specific set of generated stereotypes.

\subsection{Summary}

The main findings therefore confirm our hypothesis that performance improvements can be achieved by introducing stereotype conceptualisation annotations to the training data set, and by structuring the implied stereotypes in the \texttt{[TARGETED GROUP] [RELATION] [IMPLIED STATEMENT]} tuple format, in accordance with the CO-STAR framework. Whereas there are some instances of comparable model performances between the three different models (and indeed, this is the case for most of the 500 manually inspected model outputs from the development set), a handful of posts shows that there is a noticeable performance improvement among the models trained with the full CO-STAR framework when compared to either of the two ablated models. We observe that the -cs and -sc models are able to generate more accurate and context-specific stereotypes more often than the ablated models, and where there is a performance difference between the former two models, the -sc and -s models often have similar outputs regardless of whether or not their generated stereotypes are accurate. In any case, these three CO-STAR models exhibit a much more refined and precise set of generated stereotypes than the model trained from the work by \citet{Sap-Social}, namely SBF-GPT\textsubscript{2}.

\section{Conclusion}

\subsection{Contributions}

Understanding entailed implications and stereotypes remains a challenging task even for current state-of-the-art language models. Indeed, the subtle nature of the language commonly used to convey stereotypes, coupled with the common-sense knowledge necessary to interpret them, makes stereotype analysis and reasoning difficult for humans as well. In this paper, we present the novel CO-STAR framework for implied stereotype generation, which structures the implied statements for improved learning and evaluation, and incorporates stereotype conceptualisations in order to capture the underlying concepts of a given post. We annotate the training data set which contains 12K posts and on which the CO-STAR models are trained, and we make further improvements upon the state-of-the-art models. We manually evaluate the CO-STAR models on the development set, and show that they are able to generate more accurate and context-specific stereotypes. We also perform ablation studies to confirm the effects of the stereotype conceptualisation component of the CO-STAR framework.

\subsection{Future Work}

We note that despite the performance improvements of our CO-STAR models, they are still unable to understand more subtly worded posts and generate accurate and context-specific implied stereotypes. This necessitates further research in this multifaceted and challenging domain, studying other approaches to encode common-sense knowledge and developing more sophisticated models to understand and generate (and even neutralise) implied stereotypes.

We also acknowledge the limitations present in our newly annotated CO-STAR training data set. More specifically, the data set is susceptible to annotator bias, due to the skewed annotator demographic, as well as the fact that we were only able to assign one annotator per post, rendering any form of annotator agreement difficult to quantify. Further, the lack of annotated development and test data sets makes evaluation of the models challenging, bias-prone and time-consuming. Future work could build on our initial annotation efforts and address these limitations, and compile data sets which are more representative and complete.

\section*{Acknowledgments}

We are grateful to Ovidiu Șerban for providing constructive advice and helpful insights throughout the project. This research was funded in part by Imperial College London and its Corporate Partnership Programme.

\bibliographystyle{acl_natbib}
\bibliography{anthology,acl2021}

\appendix \label{annotation-screenshots}

\begin{figure*}
    \centering
    \includegraphics[width=0.95\linewidth]{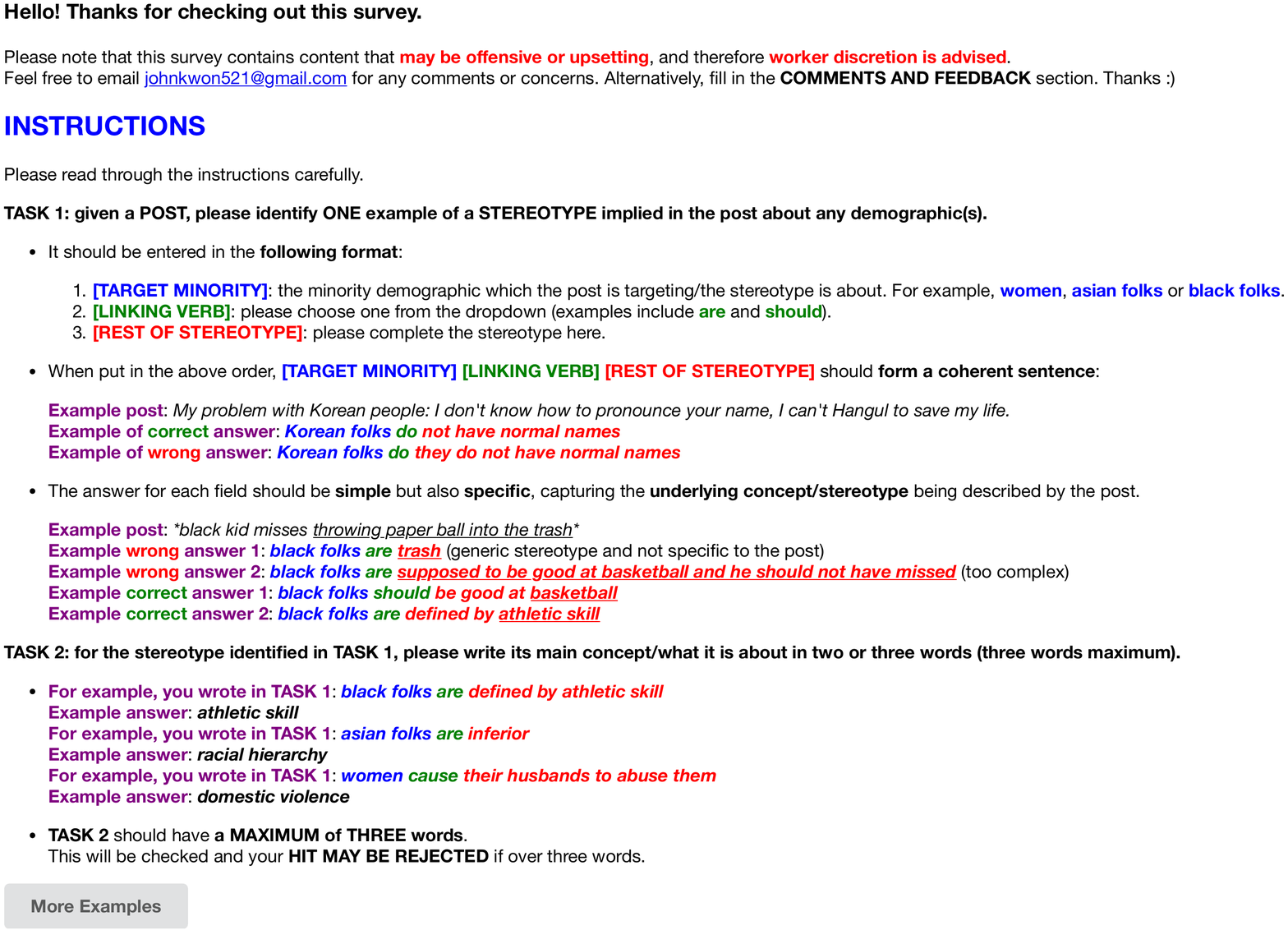}
    \includegraphics[width=0.95\linewidth]{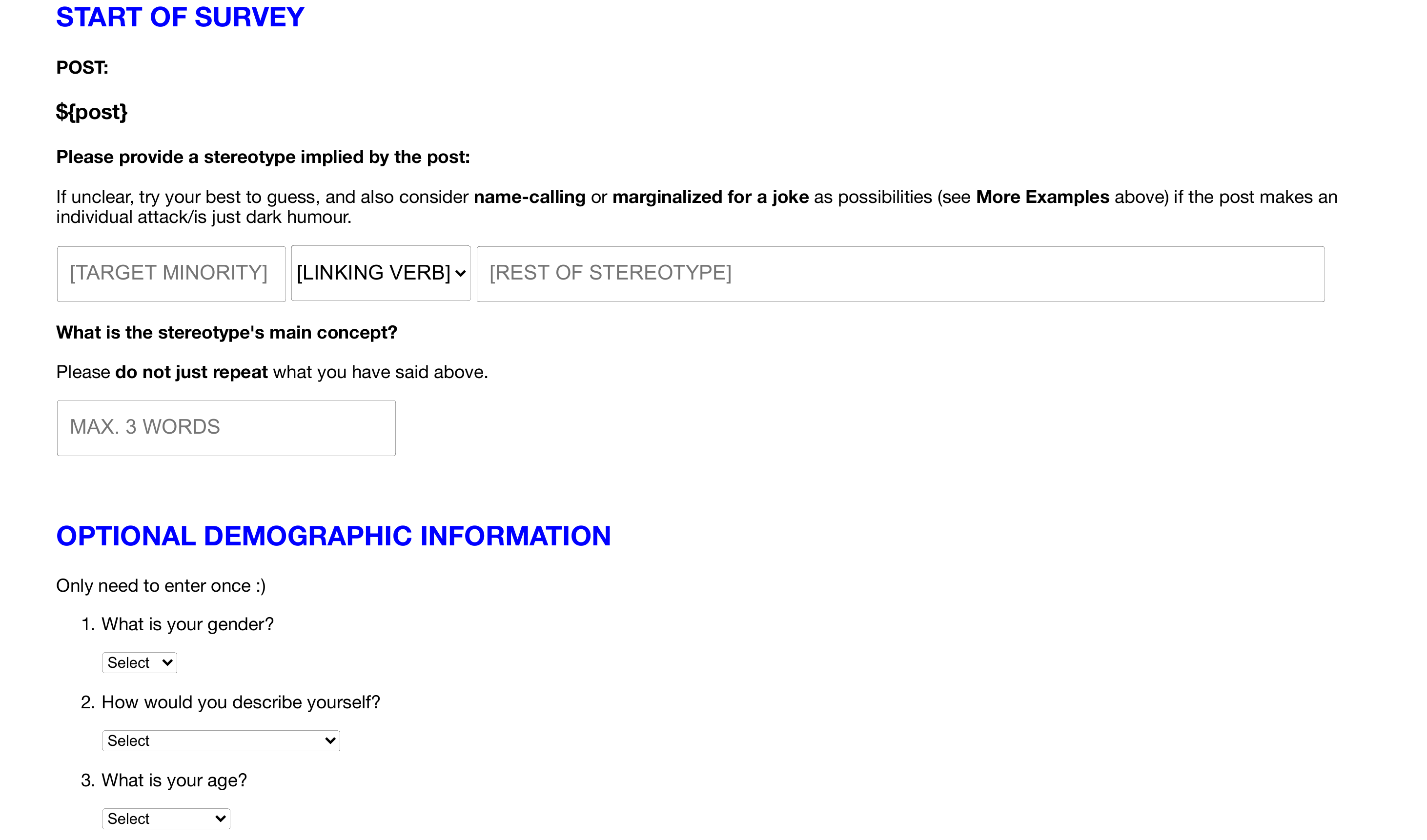}
    \includegraphics[width=0.95\linewidth]{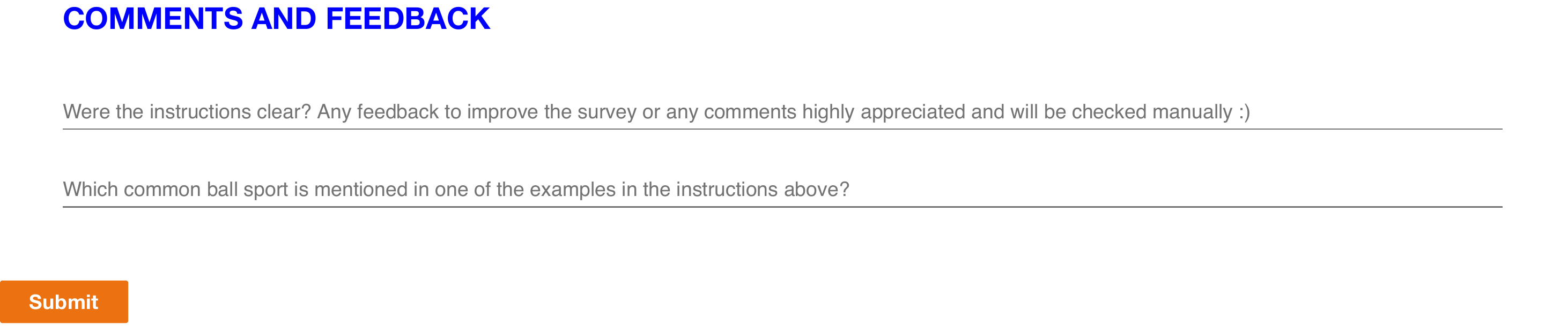}
    \caption{Screenshot of the Amazon Mechanical Turk survey.}
    \label{annotation-screenshot-1}
\end{figure*}

\end{document}